%% file: paper_template.tex
\documentclass[conference]{IEEEtran}
\usepackage{times}

\usepackage[numbers]{natbib}
\usepackage{multicol}
\usepackage[bookmarks=true]{hyperref}
\usepackage{color}
\let\labelindent\relax

\usepackage{graphicx}
\usepackage{amsmath}
\usepackage{amsfonts}       
\usepackage{
tikz,
relsize,
booktabs
}
\usepackage[font={small}]{caption}
\usepackage{bbm}
\usepackage[normalem]{ulem}
\usepackage{lipsum}

\usepackage{colortbl}
\usepackage{amssymb}
\usepackage{pifont}

\usepackage{algorithm}
\usepackage{algpseudocode}

\usepackage[normalem]{ulem}
\useunder{\uline}{\ul}{}

\usepackage[utf8]{inputenc} 
\usepackage[T1]{fontenc}    
\usepackage{url}            
\usepackage{nicefrac}       
\usepackage{microtype}      
\usepackage{xcolor}         
\usepackage{amssymb}         
\usepackage{blindtext}
\usepackage{wrapfig}
\usepackage{bbm}
\usepackage{breqn}
\usepackage{subcaption}
\usepackage{arydshln}
\usepackage{enumitem}

\usepackage{tikz}

\definecolor{olivegreen}{HTML}{3C8031}
\newcommand{\xmark}{\ding{55}}%
\newcommand{\redcross}{\textcolor{red}{\xmark}}
\newcommand{\cmark}{\ding{51}}%
\newcommand{\greencheck}{\textcolor{olivegreen}{\cmark}}

\newcommand{\xxnote}[3]{}
\ifx\hidenotes\undefined
  \renewcommand{\xxnote}[3]{}
\fi

\newcommand{\SH}[1]{{\xxnote{SH}{orange}{#1}}}

\hypersetup{
    colorlinks=true,
    linkcolor=blue,
    filecolor=magenta,      
    citecolor=blue,
    urlcolor=blue,
}

\input{math_commands.tex}

\newcommand{\method}{Point Policy} 
\newcommand{\website}{\href{https://point-policy.github.io/}{point-policy.github.io}}

\IEEEoverridecommandlockouts

\begin{document}

\title{\method{}: Unifying Observations and Actions with Key Points for Robot Manipulation} 

\author{
Siddhant Haldar\thanks{Correspondence to: siddhanthaldar@nyu.edu} \qquad Lerrel Pinto \\
\\
New York University \\
\\
{\small \tt \website{}}
}

\makeatletter
\let\@oldmaketitle\@maketitle%
\renewcommand{\@maketitle}{\@oldmaketitle%
    \centering
    \vspace{0.2in}
    \includegraphics[width=0.95\linewidth]{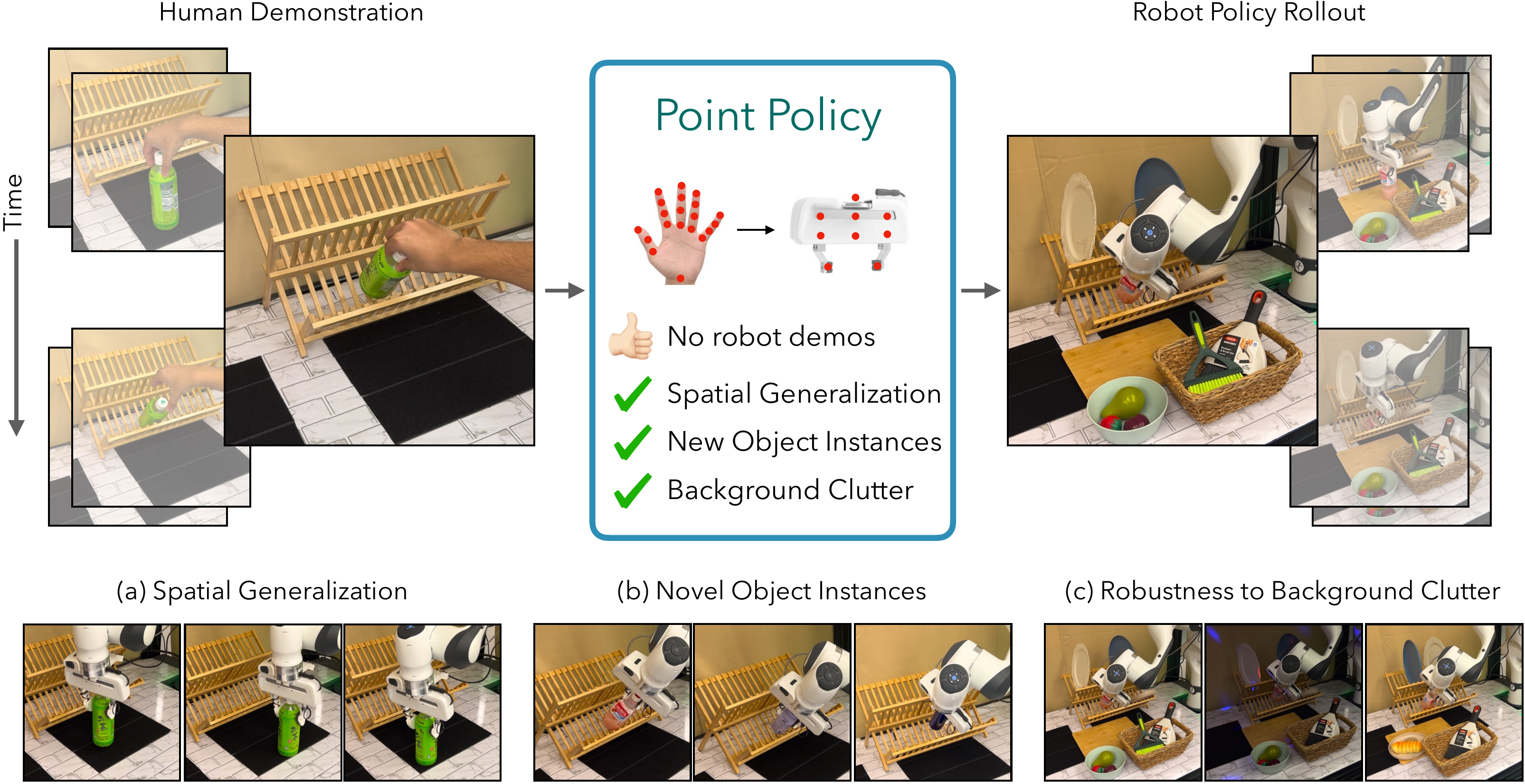}
    \captionof{figure}{We present \method{}, a framework that unifies robot observations and actions with key points and enables learning robot policies exclusively from human videos. \method{} enables learning policies with improved generalization capabilities, including spatial generalization (i.e. generalization to new locations), generalization to novel object instances, and robustness to background distractors.}
    \label{fig:intro}
}
\makeatother

\maketitle

\input{documents/abstract.tex}

\IEEEpeerreviewmaketitle

\section{Introduction}
\input{documents/introduction}

\section{Related Works}
\input{documents/related_work}

\section{Background}
\input{documents/background}

\section{\method{}}
\input{documents/approach}

\section{Experiments}
\input{documents/experiments}

\section{Conclusion and Limitations}
\input{documents/limitations}

\section{Acknowledgments}
\input{documents/acknowledgements}


\bibliographystyle{plainnat}
\bibliography{references}

\clearpage
\appendix

\input{documents/appendix.tex}

\end{document}

%% file: math_commands.tex
\usepackage{amsmath,amsfonts,bm}
\usepackage{caption}
\usepackage{subcaption}

\def\eqref#1{equation~\ref{#1}}

\def\1{\bm{1}}

\DeclareMathAlphabet{\mathsfit}{\encodingdefault}{\sfdefault}{m}{sl}
\SetMathAlphabet{\mathsfit}{bold}{\encodingdefault}{\sfdefault}{bx}{n}



%% file: documents/abstract.tex
\begin{abstract}

Building robotic agents capable of operating across diverse environments and object types remains a significant challenge, often requiring extensive data collection. This is particularly restrictive in robotics, where each data point must be physically executed in the real world. Consequently, there is a critical need for alternative data sources for robotics and frameworks that enable learning from such data. In this work, we present \method{}, a new method for learning robot policies exclusively from offline human demonstration videos and without any teleoperation data. \method{} leverages state-of-the-art vision models and policy architectures to translate human hand poses into robot poses while capturing object states through semantically meaningful key points. This approach yields a morphology-agnostic representation that facilitates effective policy learning. Our experiments on 8 real-world tasks demonstrate an overall 75\% absolute improvement over prior works when evaluated in identical settings as training. Further, \method{} exhibits a 74\% gain across tasks for novel object instances and is robust to significant background clutter. Videos of the robot are best viewed at \website{}. 
    
\end{abstract}


%% file: documents/introduction.tex
Recent years have witnessed remarkable advancements in computer vision (CV) and natural language processing (NLP), resulting in models capable of complex reasoning~\cite{gpt4,gemini,llama}, generating photorealistic images~\cite{dalle3,imagen} and videos~\cite{sora}, and even writing code~\cite{devin}. A driving force behind these breakthroughs has been the abundance of data scraped from the internet. In contrast, robotics has yet to experience a similar revolution, with most robots still confined to controlled or structured environments. While CV and NLP can readily take advantage of large-scale datasets from the internet, robotics is inherently interactive and requires physical engagement with the world for data acquisition. This makes collecting robot data significantly more challenging, both in terms of time and financial resources.

A prominent approach for training robot policies has been the collection of extensive datasets, often through contracted teleoperators~\cite{roboturk,rt1,dobbe}, followed by training deep networks on these datasets~\cite{dobbe,rums,openx,droid}. While effective, these methods tend to require months or even years of human effort~\cite{rt1,droid} and still result in datasets orders of magnitude smaller than those used in CV and NLP~\cite{openx,droid}. A potential solution to this data scarcity in robotics is to tap into the vast repository of human videos available online, showcasing individuals performing a wide range of tasks in diverse scenarios. \looseness=-1

The primary challenge in learning robot policies from human videos lies in addressing the morphology gap between robots and the human body~\cite{whirl,hudor,track2act,gen2act,motiontracks}. Two notable trends have emerged in efforts to utilize human data for learning robot policies: (1) first learning visual representations or coarse policies from human datasets and then finetuning them for downstream learning on robot datasets~\cite{track2act,gen2act,motiontracks,r3m,vptr,gr1,vip,liv,voltron}, and (2) using human videos to compute rewards for autonomous policy learning through reinforcement learning~\cite{xirl,whirl,hudor,kumar2023graph}. While the former requires a substantial amount of robot demonstrations to learn policies for downstream tasks, the latter often requires large amounts of online robot interactions in the real world, which can be time-consuming and potentially unsafe. \looseness=-1

In this work, we introduce \method{}, a new technique to learn robot policies solely from offline human data without requiring robot interactions during training. Our key observation in building \method{} is that both humans and robots occupy the same 3D space in the world, which can be tied together using key points derived from state-of-the-art vision models. 

Concretely, \method{} works in three steps. 
First, given a dataset of human videos, a motion track of key points on the human hand and the object is computed using hand pose detectors~\cite{mediapipe, hamer} and minimal human annotation of one frame per task. These key points are computed from two camera views, which allows for projection in 3D using point triangulation. Second, a transformer-based policy~\cite{baku} is trained to predict future robot points given the set of key points derived in the previous stage. Third, during inference, the predicted future robot points in 3D space are used to backtrack the 6 DOF pose of the robot's end-effector using constraints from rigid-body geometry. The gripper state of the robot end effector is predicted as an additional token. The predicted end-effector pose and gripper state are then executed on the robot at 6 Hz.

We demonstrate the effectiveness of \method{} through experiments on 8 real-world tasks on a Franka robot. Our main findings are summarized below:

\begin{enumerate}[leftmargin=*,align=left]
    \item  \method{} exhibits an absolute improvement of 75\% over prior state-of-the-art policy learning algorithms across 8 real world tasks when evaluated in identical settings as training. (Section~\ref{subsec:spatial_gen}). 
    \item \method{} generalizes to novel object instances, exhibited a 74\% absolute improvement over prior work on a held-out set of objects unseen in the training data. (Section~\ref{subsec:novel_gen}). 
    \item Policies trained with \method{} are robust to the presence of background distractors, performing at par with scenes without clutter (Section~\ref{subsec:distractor}). 
    \item We provide an analysis of co-training \method{} with teleoperated robot data (Section~\ref{subsec:robot_data}) and study the importance of several design choices in \method{} (Section~\ref{subsec:design_considerations}).
\end{enumerate}


All of our datasets, and training and evaluation code have been made publicly available. Videos of our
trained policies can be seen here: \website{}.

%% file: documents/related_work.tex
\subsection{Imitation Learning}
Imitation Learning (IL)~\cite{imitation_learning} refers to training policies with expert demonstrations, without requiring a predefined reward function. In the context of reinforcement learning (RL), this is often referred to as inverse RL~\cite{irl_1, irl_2}, where the reward function is derived from the demonstrations and used to train a policy~\cite{wayex, rot, fish, gail, nair2020awac}. While these methods reduce the need for extensive human demonstrations, they still suffer from significant sample inefficiency. As a result of this inefficiency in deploying RL policies in the real world, behavior cloning (BC)~\cite{Pomerleau-1989-15721, torabi2019recent, schaal1996learning, ross2011reduction} has become increasingly popular in robotics. Recent advances in BC have demonstrated success in learning policies for both long-horizon tasks~\cite{sequential_dexterity,learning_to_generalize,clipport} and multi-task scenarios~\cite{baku,roboagent,rtx,track2act,gen2act}. However, most of these approaches rely on image-based representations~\cite{zhang2018deep,baku,diffusionpolicy,roboagent,rtx,bcz}, which limits their ability to generalize to new objects and function effectively outside of controlled lab environments. In this work, we propose \method{}, which attempts to address this reliance on image representations by directly using key points as an input to the policy instead of raw images. Through extensive experiments, we observe that such an abstraction helps learn robust policies that generalize across varying scenarios.\looseness=-1

\subsection{Object-centric Representation Learning} 
Object-centric representation learning aims to create structured representations for individual components within a scene, rather than treating the scene as a whole. Common techniques in this area include segmenting scenes into bounding boxes~\cite{deep_object_centric, learning_to_generalize, one_shot, fang2023anygrasp, viola} and estimating object poses~\cite{deep_object_pose_estimation, hope}. While bounding boxes show promise, they share similar limitations with non object-centric image-based models, such as overfitting to specific object instances. Pose estimation, although less prone to overfitting, requires separate models for each object in a task. Another popular method involves using point clouds~\cite{groot, reagentpointcloudregistration}, but their high dimensionality necessitates specialized models, making it difficult to accurately capture spatial relationships. Lately, several works have resorted to adopting key points~\cite{p3po,ju2025robo,rekep,track2act,gen2act,motiontracks,fang2024keypoint,bechtle2023multimodal} for policy learning due to their generalization ability. Further, key points also allow the direct injection of human priors into the policy learning pipeline~\cite{track2act,gen2act,motiontracks} as opposed to learning representations from human videos followed by downstream learning on robot teleoperated data~\cite{r3m,vptr,gr1,vip,liv,voltron}. In this work, we leverage key points as a unified observation and action space to enable learning generalizable policies exclusively from human videos. \looseness=-1

\setcounter{figure}{1} 
\begin{figure*}[t]
\centering
\includegraphics[width=\linewidth]{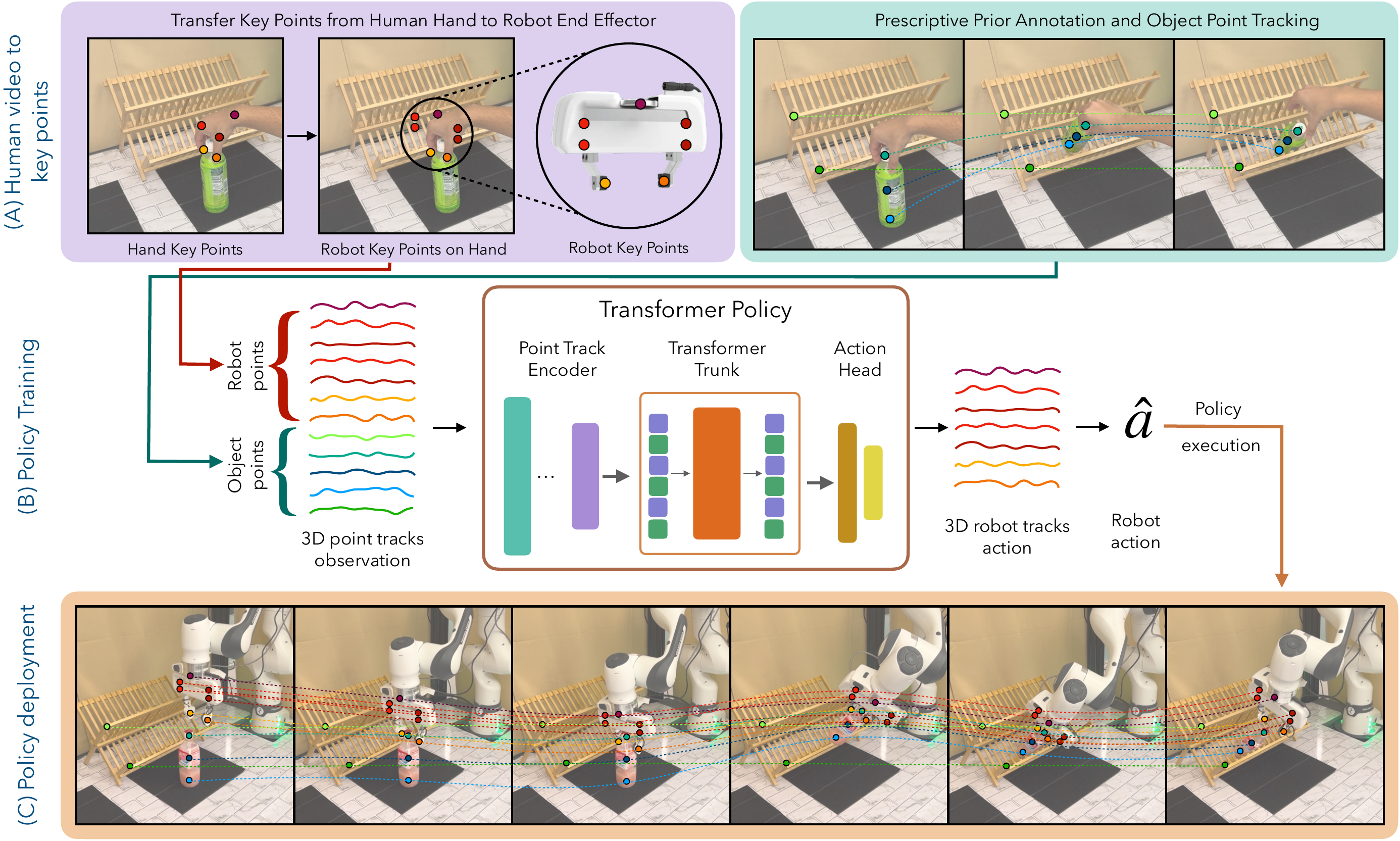}
\vspace{-1em}
\caption{Overview of the \method{} framework. (a) \method{} leverages state-of-the-art vision models and policy architectures to translate human hand poses into robot poses while capturing object states through sparse single-frame human annotations. (b) The derived key points are fed into a transformer policy to predict the 3D future point tracks from which the robot actions are computed through rigid-body geometry constraints. (c) Finally, the computed action is executed on the robot using end-effector position control at a 6Hz frequency.}
\label{fig:method}
\end{figure*}

\subsection{Human-to-Robot Transfer for Policy Learning}
There have been several attempts at learning robot policies from human videos. Some works first learn visual representations from large-scale human video datasets and learn a downstream policy on these representations using limited amounts of robot data~\cite{r3m,vptr,gr1,vip,liv,voltron}. Another line of work learns coarse policies from human videos, using key points~\cite{track2act} and generative modeling~\cite{gen2act}, which are then improved using downstream learning on robot data. Recently proposed MT-$\pi$~\cite{motiontracks} alleviates the need for downstream learning by co-training a key point policy with human and robot data. A caveat in all these works is that despite having access to abundant human demonstrations, there is a need to collect robot data to achieve a highly performant policy. A recently emerging line of work~\cite{r+x} attempts to do away with this need for robot data by doing in-context learning with state-of-the-art vision-language models (VLMs)~\cite{gemini,gpt4,llama}. However, owing to the large compute times of VLMs, these policies are required to be deployed open-loop and hence, are not reactive to changes in the scene. In this work, we propose \method{}, a new framework that learns generalizable policies from human videos, does not require robot demonstrations or online robot interactions, and can be executed in a closed-loop fashion. \looseness=-1

%% file: documents/background.tex
\subsection{Imitation learning}
The goal of imitation learning is to learn a behavior policy $\pi^b$ given access to either the expert policy $\pi^e$ or trajectories derived from the expert policy $\tau^e$. This work operates in the setting where the agent only has access to observation-based trajectories, i.e. $\tau^e \equiv \{(o_t, a_t)_{t=0}^{T}\}_{n=0}^N$. Here $N$ and $T$ denote the number of demonstrations and episode timesteps respectively. We choose this specific setting since obtaining observations and actions from expert or near-expert demonstrators is feasible in real-world settings~\cite{aloha,openteach} and falls in line with recent work in this area~\cite{baku,vqbet,aloha,diffusionpolicy}.

\subsection{Behavior Cloning}
Behavior Cloning (BC)~\cite{pomerleau1998autonomous,shafiullah2024supervised} corresponds to solving the maximum likelihood problem shown in Eq.~\ref{eq:bc}. Here $\mathcal{T}^{e}$ refers to expert demonstrations. When parameterized by a normal distribution with fixed variance, the objective can be framed as a regression problem where, given observations $o^e$, $\pi^{BC}$ needs to output $a^e$.
\begin{equation}
    \mathcal{L}^{BC} = \mathbb{E}_{(o^{e},a^{e})\sim \mathcal{T}^{e}} \|a^{e} - \pi^{BC}(o^{e})\|^{2}
    \label{eq:bc}
\end{equation}

 After training, it enables $\pi^{BC}$ to mimic the actions corresponding to the observations seen in the demonstrations.

\subsection{Semantic Correspondence and Point Tracking}
Semantic correspondence and point tracking are fundamental problems in computer vision. Semantic correspondence matches semantically equivalent points between images of different scenes, while point tracking follows reference points across video frames. We leverage these ideas using two state-of-the-art models: DIFT~\cite{dift} and Co-Tracker~\cite{cotracker}. DIFT establishes correspondences between reference and observed images, as illustrated in Figure~\ref{fig:correspondence}, while Co-Tracker tracks initialized key points throughout the video trajectory (Figure~\ref{fig:method}). This integration enables robust identification and tracking of semantically meaningful points across diverse visual scenarios, forming a key component \method{}. We have included a more detailed explanation in Appendix~\ref{appendix:background}.

%% file: documents/approach.tex
\method{} seeks to learn generalizable policies exclusively from human videos that are robust to significant environmental perturbations and applicable to diverse object locations and types. An overview of our method is presented in Figure~\ref{fig:method}. Before diving into the details, we first present some of the key assumptions needed to run \method{}.

\textbf{Assumptions:} (1) The pose of the human hand in the first frame is known for each task. This is needed to initialize the robot and set that pose as the base frame of operation. This assumption can be relaxed with a hand-pose estimator~\cite{hamer}, which we do not investigate in this work. (2) We operate in a calibrated scene with the camera's intrinsic and extrinsic matrices, and the transforms between each camera and the robot base known. In practice this is a one-time process that takes under 5 minutes when the robot system is first installed.

\subsection{Point-based Scene Representation}
\label{subsec:scene_rep}

Our method begins by collecting human demonstrations, which are then converted to a point-based representation amenable to policy learning. 

\subsubsection{Human-to-Robot Pose Transfer}
For each time step $t$ of a human video, we first extract image key points on the human hand $p_h^t$ using the MediaPipe~\cite{mediapipe} hand pose detector, focusing specifically on the index finger and thumb. The corresponding hand key points $p_h^t$ obtained from two camera views are used to compute the 3D world coordinates $\mathcal{P}_h^t$ of the human hand through point triangulation. We use point triangulation for 3D projection due to its higher accuracy as compared to sensor depth from the camera (Section~\ref{subsec:design_considerations}). The robot position $\mathcal{R}_{pos}^t$ is computed as the midpoint between the tips of the index finger and thumb in $\mathcal{P}_h^t$. The robot orientation $\mathcal{R}_{ori}^t$ is computed as

\begin{equation}
    \label{eq:orientation}
    \begin{aligned}
        \Delta \mathcal{R}_{ori}^t &= \mathcal{T}(\mathcal{P}_h^{0}, \mathcal{P}_h^{t}) \\
        \mathcal{R}_{ori}^t &= \Delta \mathcal{R}_{ori}^t \cdot \mathcal{R}_{ori}^{0}
    \end{aligned}
\end{equation}

where $\mathcal{T}$ computes the rigid transform between hand key points on the first frame of the video, $\mathcal{P}_h^{0}$, and $\mathcal{P}_h^{t}$. The robot end effector pose is then represented at $T_r^t \leftarrow\{\mathcal{R}_{pos}^t, \mathcal{R}_{ori}^t\}$. The robot's gripper state $\mathcal{R}_g$ is computed using the distance between the tip of the index finger and thumb. The gripper is considered closed when the distance is less than 7cm, otherwise open. 
Finally, given the robot pose $T_r^t$, we define a set of $N$ rigid transformations $T$ about the computed robot pose and compute robot key points $\mathcal{P}_r^t$ such that

\begin{equation}
    (\mathcal{P}_r^t)^i = T_r^t \cdot T^i, ~~\forall i \in \{1, ..., N\}
\end{equation}

This process has been demonstrated in Figure~\ref{fig:method}. This approach effectively bridges the morphological gap between human hands and robot manipulators, enabling accurate transfer of demonstrated actions to a robotic framework.

\subsubsection{Environment state through point priors}
To obtain key points on task-relevant objects in the scene, we adopt the method proposed by P3PO~\cite{p3po}. Initially, a user randomly selects one demonstration from a dataset of human videos and annotates semantically meaningful object points on the first frame that are pertinent to the task being performed. This annotation process is quick, taking only a few seconds. The user-annotated points serve as priors for subsequent data generation. Using an off-the-shelf semantic correspondence model, DIFT~\cite{dift}, we transfer the annotated points from the first frame to the corresponding locations in the first frames of all other demonstrations within the dataset. This approach allows us to initialize key points throughout the data set with minimal additional human effort. 

\begin{figure}[t]
\centering
\includegraphics[width=\linewidth]{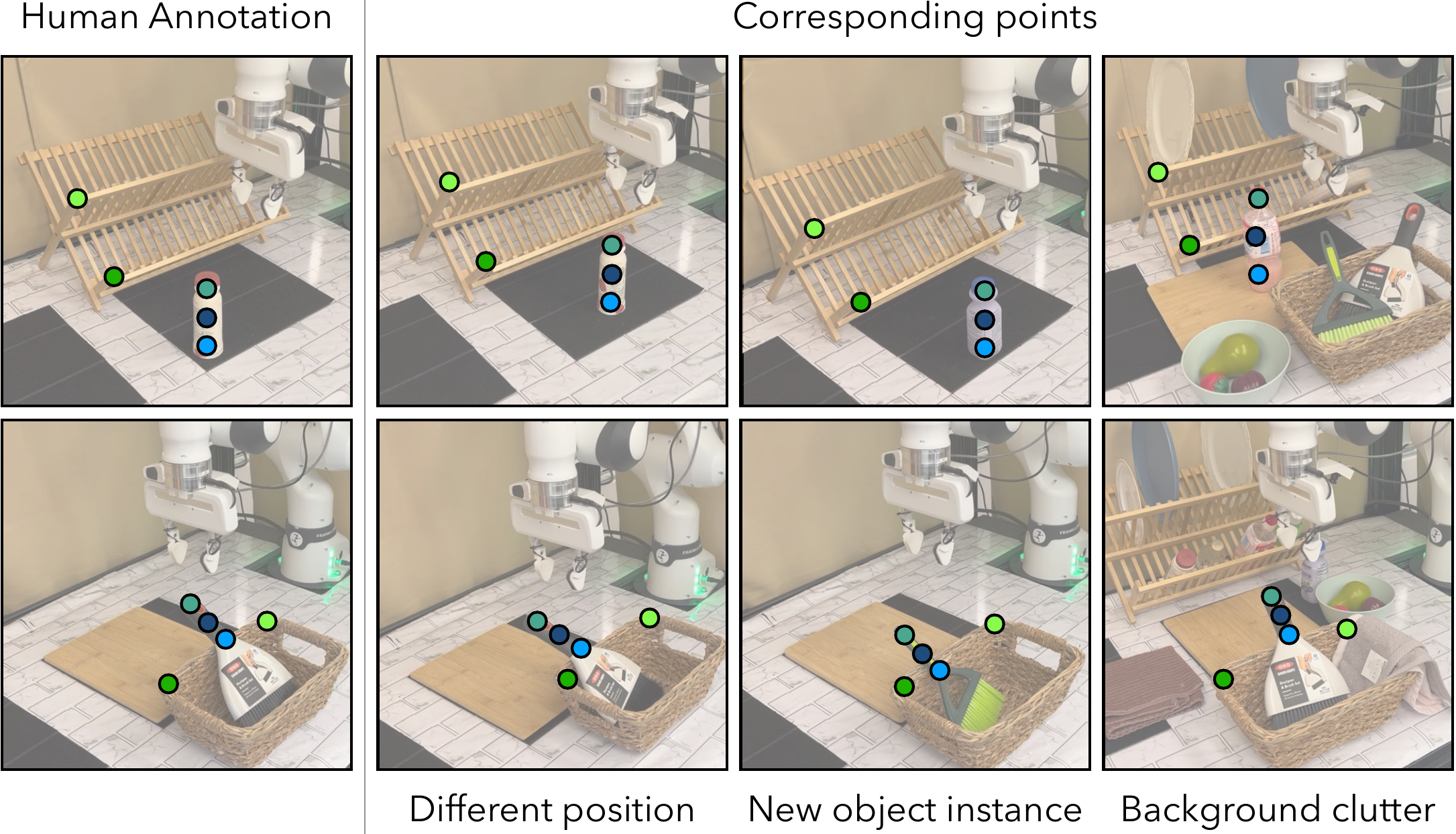}
\caption{Results of the correspondence model when used for the put bottle on rack and sweep broom tasks. On the left is a frame with human annotations for the object points. On the right, we show that semantic correspondence can identify the same points across different positions, new object instances, and background clutter.}
\vspace{-1em}
\label{fig:correspondence}
\end{figure}

For each demonstration, we then employ Co-Tracker~\cite{cotracker}, an off-the-shelf point tracker, to automatically track these initialized key points throughout the entire trajectory. By leveraging existing vision models for correspondence and tracking, we efficiently compute object key points for every frame in the dataset while requiring user input for only a single frame. This process, illustrated in Figure~\ref{fig:correspondence}, capitalizes on large-scale pre-training of vision models to generalize across new object instances and scenes without necessitating further training. We prefer point tracking over correspondence at each frame due to its faster inference speed and its capability to handle occlusions by continuing to track points. The corresponding object points from two camera views are lifted to 3D world coordinates using point triangulation to obtain the 3D object key points $\mathcal{P}_o$. During inference, DIFT is employed to identify corresponding object key points on the first frame, followed by Co-Tracker tracking these points during execution. \looseness=-1

It is important to note that \method{} utilizes multiple camera views only for point triangulation, with the policy being learned on 3D key points grounded in the robot's base frame. More details on point triangulation can be found in Appendix~\ref{appendix:point_triangulation}. \looseness=-1

\subsection{Policy Learning}
\label{subsec:policy_learning}
For policy learning, we use BAKU~\cite{baku}. Instead of providing raw images as input, we provide the robot points $\mathcal{P}_r$ and object points $\mathcal{P}_o$ grounded in the robot's base frame as input to the policy. A history of observations for each key point is flattened into a single vector which is then encoded using a multilayer perceptron (MLP) encoder. The encoded representations are fed as separate tokens along with a gripper token into a BAKU~\cite{baku} transformer policy, which predicts the future tracks for each robot point $\mathcal{\hat P}_r$ and the robot gripper state $\mathcal{\hat G}_r$ using a deterministic action head. Mathematically, this can be represented as 

\begin{equation}
    \begin{aligned}
        \mathcal{O}^{t-H:t} &= \{\mathcal{P}_r^{t-H:t},~\mathcal{P}_o^{t-H:t}\}\\
        \mathcal{\hat P}_r^{t+1},~\mathcal{G}_r^{t+1} &= \pi(\cdot|~\mathcal{O}^{t-H:t})
    \end{aligned}
\end{equation}

where $H$ is the history length and $\pi$ is the learned policy. Following prior works in policy learning~\cite{aloha, diffusionpolicy}, we use action chunking with exponential temporal averaging to ensure temporal smoothness of the predicted point tracks. The transformer is non-causal in this scenario and hence the training loss is only applied to the robot point tracks.

\subsection{Backtrack Robot Actions from Predicted Key Points}
\label{subsec:robot_action_from_points}
The predicted robot points $\mathcal{\hat P}_r$ are mapped back to the robot pose using constraints from rigid-body geometry. We first consider the key point corresponding to the robot's wrist $\mathcal{\hat P}_r^{wrist}$ as the robot position $\mathcal{\hat R}_{pos}$. The robot orientation $\mathcal{\hat R}_{ori}$ is computed using Eq.~\ref{eq:orientation} considering $\mathcal{R}_{ori}^{0}$ is fixed and known. Finally, the robot action $\mathcal{A}_r$ is defined as 

\begin{equation}
    \mathcal{\hat A}_r = (\mathcal{\hat R}_{pos},~\mathcal{\hat R}_{ori},~\mathcal{\hat G}_r)
\end{equation}

Finally, the action $\mathcal{\hat A}_r$ is executed on the robot using end-effector position control at a 6Hz frequency.

%% file: documents/experiments.tex
Our experiments are designed to answer the following questions: $(1)$ How well does \method{} work for policy learning? $(2)$ How well does \method{} work for novel object instances? $(3)$ Can \method{} handle background distractors? $(4)$ Can \method{} be improved with robot demonstrations? $(5)$ What design choices matter for human-to-robot learning?

\subsection{Experimental Setup}
Our experiments utilize a Franka Research 3 robot equipped with a Franka Hand gripper, operating in a real-world environment. We use the Deoxys~\cite{viola} real-time controller for controlling the robot. The policies utilize RGB and RGB-D images captured using Intel RealSense D435 cameras from two third-person camera views. The action space encompasses the robot's end effector pose and gripper state. We collect a total of 190 human demonstrations across 8 real-world tasks, featuring diverse object positions and types. Additionally, for studying the effect of co-training with robot data (Section~\ref{subsec:robot_data}), we collect a total of 100 robot demonstrations for 4 tasks (Section~\ref{subsec:robot_data}) using a VR-based teleoperation framework~\cite{openteach}. All demonstrations are recorded at a 20Hz frequency and subsequently subsampled to approximately 6Hz. For methods that directly predict robot actions, we employ absolute actions during training, with orientation represented using a 6D rotation representation~\cite{zhou2019continuity}. This representation is chosen for its continuity and fast convergence properties. The learned policies are deployed at a 6Hz frequency during execution. \looseness=-1

\begin{figure*}[t]
\centering
\includegraphics[width=0.9\linewidth]{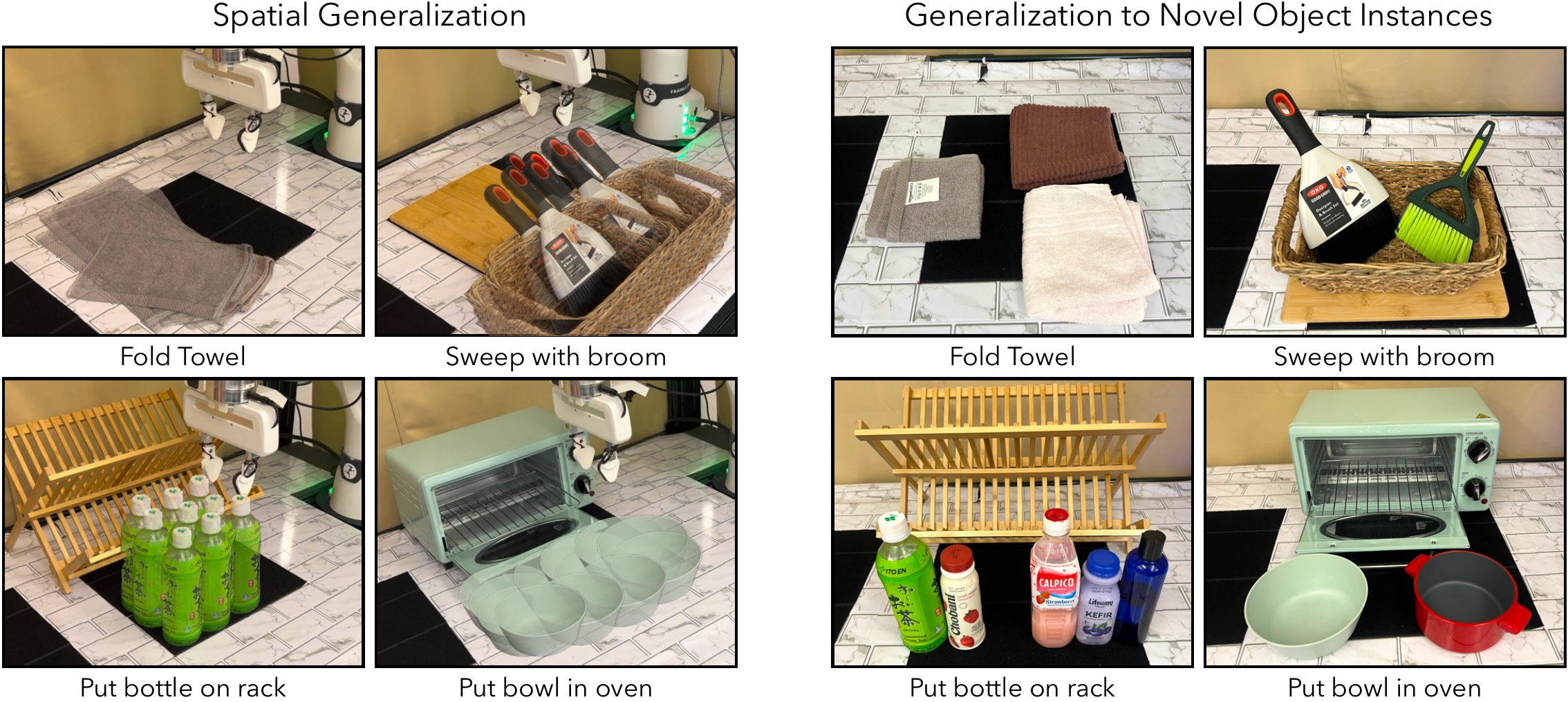}
\caption{\textbf{(left)} Illustration of spatial variation used in our experiments. \textbf{(right)} Range of objects used in our experiments, where the objects on the left are in-domain objects while on the right are unseen objects used in our generalization experiments.}
\vspace{-1em}
\label{fig:generalization}
\end{figure*}

\subsection{Task Descriptions}
\label{subsec:task_desc}
\begin{figure*}[t]
\centering
\includegraphics[width=\linewidth]{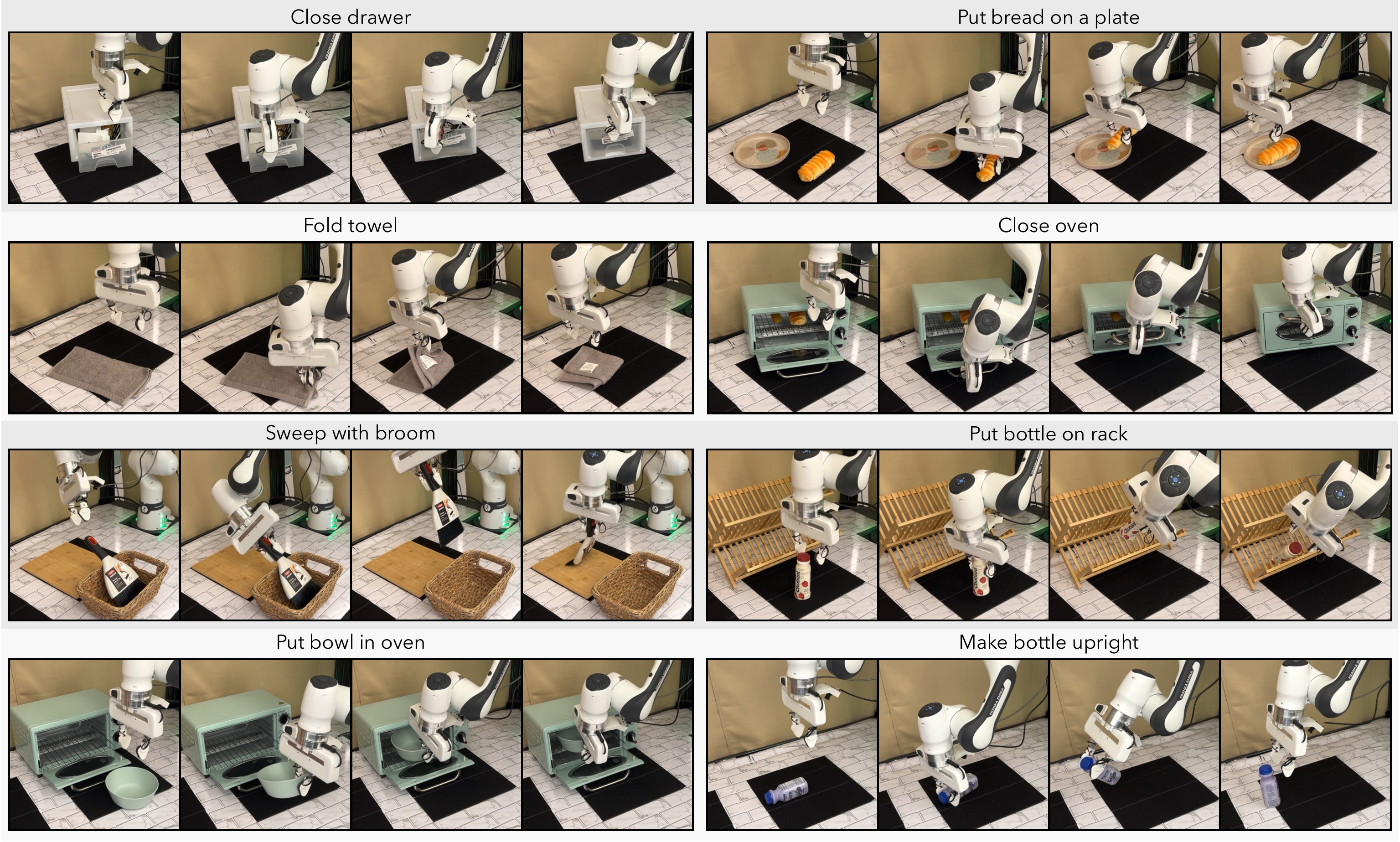}
\caption{Real-world rollouts showing \method{}'s ability on in-domain objects across 8 real-world tasks.}
\label{fig:rollouts}
\end{figure*}

We experiment with manipulation tasks with significant variability in object position, type, and background context. Figure~\ref{fig:rollouts} depicts rollouts for all of our tasks. For each task, we collect data across various object sizes and appearances. During evaluations, we add novel object instances that are unseen during training. The variations in positions and object instances for selected tasks are depicted in Figure~\ref{fig:generalization}, with more examples provided in Appendix~\ref{appendix:novel_objects}. We provide a brief description of each task below.

\paragraph{Close drawer} The robot arm is tasked with pushing close a drawer placed on the table. The position of the drawer varies for each evaluation. We collect 20 demonstrations for a single drawer and run evaluations on the same drawer.

\paragraph{Put bread on plate} The robot arm picks up a piece of bread from the table and places it on a plate. The positions of the bread and the plate are varied for each evaluation. We collect 30 demonstrations for the task of a single bread-plate pair. During evaluations, we introduce two new plates.

\paragraph{Fold towel} The robot arm picks up a towel placed on the table from a corner and folds it. The position of the towel varies for each evaluation. We collect 20 demonstrations for a single towel. During evaluations, we introduce two new towels. \looseness=-1

\paragraph{Close oven} The robot arm is tasked with closing the door of an oven. The position of the oven varies for each evaluation. We collect 20 demonstrations for the task on a single oven and run evaluations on the same oven.

\paragraph{Sweep broom} The robot arm picks up a broom and sweeps the table. The position and orientation of the broom are varied across evaluations. We collect 20 demonstrations for a single broom. During evaluations, we introduce a new broom. \looseness=-1

\paragraph{Put bottle on rack} The robot arm picks up a bottle from the table and places it on the lower level of a kitchen rack. The position of the bottle is varied for each evaluation. We collect 15 demonstrations for 2 different bottles, resulting in a total of 30 demonstrations for the task. During evaluations, we introduce three new bottles.

\paragraph{Put bowl in oven} The robot arm picks up a bowl from the table and places it inside an oven. The position of the bowl varies for each evaluation. We collect 20 demonstrations for the task with a single bowl. During evaluations, we introduce a new bowl.

\paragraph{Make bottle upright} The robot arm pick up a bottle from the table and places it in an upright position. The position of the bottle varies for each evaluation. We collect 15 demonstrations for 2 different bottles, resulting in a total of 30 demonstrations for the task. During evaluations, we introduce two new bottles.

\subsection{Baselines}
We compare \method{} with 4 baselines - \textit{behavior cloning (BC)}~\cite{baku} with RGB and RGB-D images, \textit{Motion Tracks}~\cite{motiontracks}, and \textit{P3-PO}~\cite{p3po}. We describe each method below.

\paragraph{Behavior Cloning (BC)~\cite{baku}} This method performs behavior cloning (BC) using the BAKU policy learning architecture~\cite{baku}, which takes RGB images of the human hand as input and predicts the extracted robot actions as output.

\paragraph{Behavior Cloning (BC) with Depth} This is similar to BC but uses both RGB and depth images as input.

\paragraph{Motion Track Policy (MT-$\pi$)~\cite{motiontracks}} Given an image of the scene and robot key points on the image, MT-$\pi$ predicts the future 2D robot point tracks to complete a task. This approach generates future 2D point tracks for robot points across multiple views, which are then triangulated to obtain 3D points on the robot. These 3D points are subsequently converted to the robot's absolute pose (similar to our proposed method) and treated as the robot's action. Implementation details for MT-$\pi$ have been provided in Appendix~\ref{appendix:motion_tracks}.

\paragraph{P3-PO~\cite{p3po}} This method utilizes image points representing both the robot and objects of interest, projecting them into 3D space using camera depth information. These 3D points serve as input to a transformer policy~\cite{baku}, which predicts robot actions. P3PO's 3D point representations, akin to those in \method{}, enable spatial generalization, adaptability to novel object instances, and robustness to background clutter.

\begin{table*}[t]
\centering
\caption{Policy performance of \method{} on in-domain object instances on 8 real-world tasks.}
\label{table:same_distribution}
\renewcommand{\tabcolsep}{4pt}
\renewcommand{\arraystretch}{1}
\begin{tabular}{lcccccccc}
\toprule
\multicolumn{1}{c}{\textbf{Method}} & \textbf{\begin{tabular}[c]{@{}c@{}}Close\\ drawer\end{tabular}} & \textbf{\begin{tabular}[c]{@{}c@{}}Put bread \\ on plate\end{tabular}} & \textbf{\begin{tabular}[c]{@{}c@{}}Fold\\ towel\end{tabular}}& \textbf{\begin{tabular}[c]{@{}c@{}}Close\\ oven\end{tabular}} & \textbf{\begin{tabular}[c]{@{}c@{}}Sweep\\ broom\end{tabular}} & \textbf{\begin{tabular}[c]{@{}c@{}}Put bottle\\ on rack\end{tabular}} & \textbf{\begin{tabular}[c]{@{}c@{}}Put bowl\\ in oven\end{tabular}} & \textbf{\begin{tabular}[c]{@{}c@{}}Make bottle\\ upright\end{tabular}} \\ \bottomrule
BC~\cite{baku}                    & 0/10 & 0/20 & 0/10 & 0/10 & 0/10 & 0/30 & 1/10 & 0/20 \\
BC w/ Depth                       & 0/10 & 0/20 & 0/10 & 0/10 & 0/10 & 0/30 & 0/10 & 0/20 \\
MT-$\pi$~\cite{motiontracks}      & 2/10 & 2/20 & 0/10 & 4/10 & 0/10 & 8/30 & 0/10 & 0/20 \\
P3-PO~\cite{p3po}                 & 0/10 & 0/20 & 0/10 & 0/10 & 0/10 & 0/30 & 0/10 & 0/20 \\
\hdashline\noalign{\vskip 0.5ex}
\method{} \textbf{(Ours)}         & \textbf{10/10} & \textbf{19/20} & \textbf{9/10} & \textbf{9/10} & \textbf{9/10} & \textbf{26/30} & \textbf{8/10} & \textbf{16/20} \\ \bottomrule
\end{tabular}
\end{table*}

\begin{table*}[t]
\centering
\renewcommand{\tabcolsep}{4pt}
\renewcommand{\arraystretch}{1}
\caption{Policy performance of \method{} on novel object instances on 6 real-world tasks.}
\label{table:novel_objects}
\begin{tabular}{lcccccc}
\toprule
\multicolumn{1}{c}{\textbf{Method}} & \textbf{\begin{tabular}[c]{@{}c@{}}Put bread \\ on plate\end{tabular}} & \textbf{\begin{tabular}[c]{@{}c@{}}Fold\\ towel\end{tabular}} & \textbf{\begin{tabular}[c]{@{}c@{}}Sweep\\ broom\end{tabular}} &  \textbf{\begin{tabular}[c]{@{}c@{}}Put bottle\\ on rack\end{tabular}} & \textbf{\begin{tabular}[c]{@{}c@{}}Put bowl\\ in oven\end{tabular}} & \textbf{\begin{tabular}[c]{@{}c@{}}Make bottle\\ upright\end{tabular}} \\ \midrule
BC~\cite{baku}               & 0/20 & 0/20 & 0/10 & 0/30 & 0/10 & 0/20 \\
BC w/ Depth                  & 0/20 & 0/20 & 0/20 & 0/30 & 0/10 & 0/20 \\
MT-$\pi$~\cite{motiontracks} & 1/20 & 0/20 & 0/10 & 0/30 & 0/10 & 0/20 \\
P3-PO~\cite{p3po}            & 0/20 & 0/20 & 0/10 & 0/30 & 0/10 & 0/20 \\
\hdashline\noalign{\vskip 0.5ex}
\method{} \textbf{(Ours)}    & \textbf{18/20} & \textbf{15/20} & \textbf{4/10} & \textbf{27/30} & \textbf{9/10} & \textbf{9/20} \\ \bottomrule
\end{tabular}
\end{table*}

\begin{table*}[t]
\centering
\renewcommand{\tabcolsep}{4pt}
\renewcommand{\arraystretch}{1}
\caption{Policy performance of \method{} with background distractors on both in-domain and novel object instances.}
\label{table:distractors}
\begin{tabular}{ccccccc}
\toprule
\multicolumn{1}{c}{\textbf{Background distractors}} &
  \multicolumn{2}{c}{\textbf{Put bread on plate}} &
  \multicolumn{2}{c}{\textbf{Sweep broom}} &
  \multicolumn{2}{c}{\textbf{Put bottle on rack}} \\
\cmidrule(lr){2-3}
\cmidrule(lr){4-5}
\cmidrule(lr){6-7}            & In-domain & Novel object & In-domain & Novel object & In-domain & Novel object \\ \midrule
\redcross                     & 19/20     & 18/20        & 9/10      & 4/10        & 26/30     & 27/30         \\
\greencheck                   & 18/20     & 18/20        & 9/10      & 2/10        & 23/30     & 23/30         \\ \bottomrule
\end{tabular}
\end{table*}

\begin{figure*}[t]
\centering
\includegraphics[width=\linewidth]{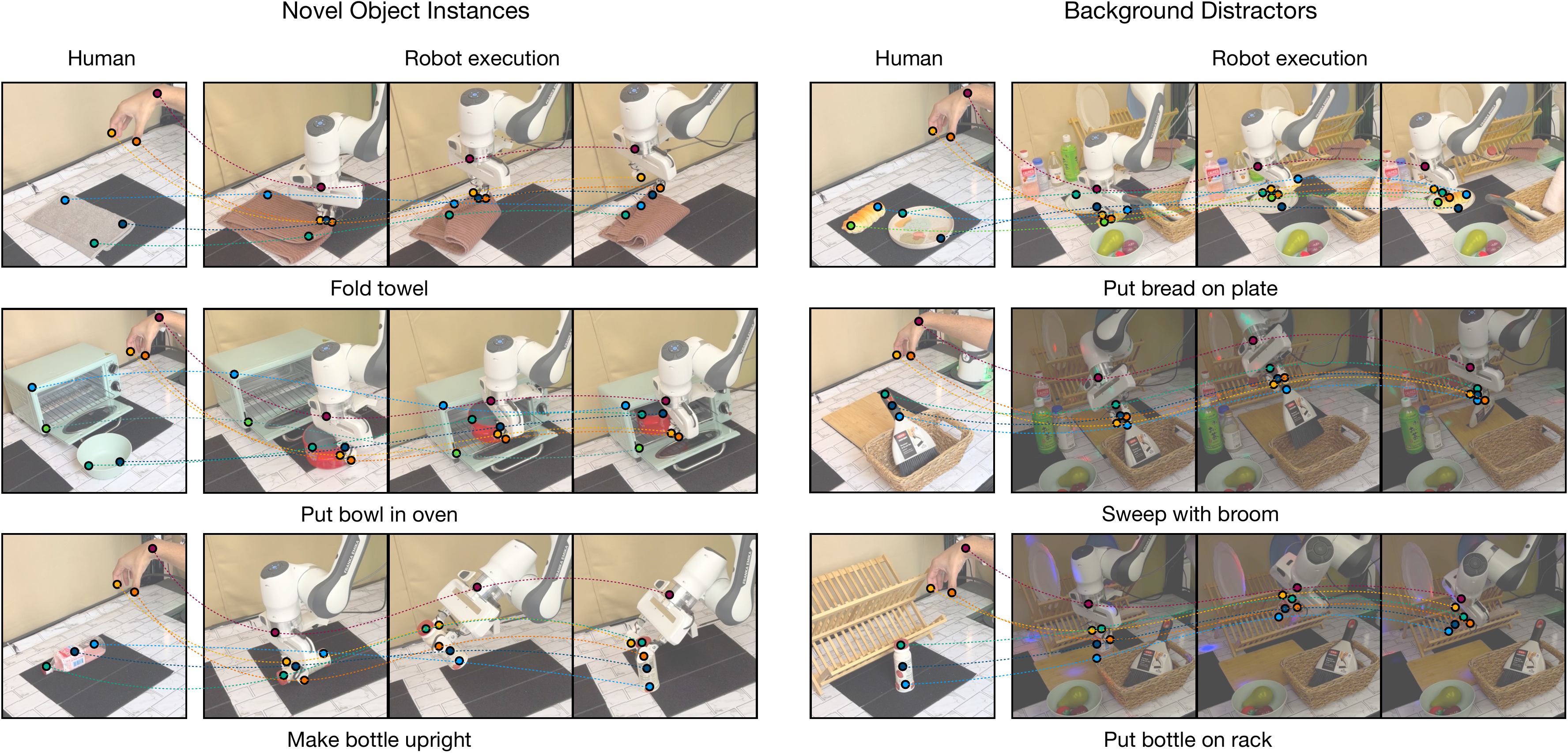}
\caption{Real-world rollouts showing that \method{} generalizes to novel object instances and is robust to background distractors.}
\label{fig:rollouts_variations}
\end{figure*}

\subsection{Considerations for policy learning}
\method{} and P3PO use a point-based representation obtained from $640\times480$ images. For correspondence, we use DIFT~\cite{dift} using the first layer of the hundredth diffusion time step with an ensemble size of 8. Point tracking is performed using a modified version of Co-Tracker~\cite{cotracker} that enables tracking one frame at a time, rather than chunks. \method{}, MT-$\pi$, and P3PO use a history of 10 point observations, while the image-based baselines do not use history~\cite{baku}. BC (RGB), BC (RGB-D), and MT-$\pi$ are trained on images of size $256\times256$. All methods predict an action chunk~\cite{aloha} of size 20 (\(\sim \) 3 seconds). \looseness=-1

\subsection{How well does \method{} work for policy learning?}
\label{subsec:spatial_gen}

We evaluate \method{} in an in-domain setting, using the same objects seen during training. The evaluation consists of 10 trials per object for each task, resulting in a variable total number of trials per task. The results of this evaluation are summarized in Table~\ref{table:same_distribution}. Baselines that rely on RGB images as inputs (RGB, RGB-D, MT-$\pi$) perform poorly when trained exclusively on human hand videos. This is largely due to the significant visual differences between the human hand and the robot manipulator. While appearance-agnostic, P3-PO struggles due to noisy depth data from the camera. 
\method{} achieves an average success rate of 88\% across all tasks, outperforming the strongest baseline MT-$\pi$ by 75\%. Overall, these results demonstrate that \method{}'s ability to effectively address challenges related to visual differences and noisy depth data, achieving state-of-the-art performance in an in-domain setting.

\subsection{How well does \method{} work for novel object instances?}
\label{subsec:novel_gen}
Table~\ref{table:novel_objects} compares the performance of \method{} when evaluated on new object instances unseen in the training data. We perform this comparison on a subset of our tasks. We observe that \method{} achieves an average success rate of 74\% across all tasks, outperforming the strongest baseline by 73\%. Compared to P3PO\cite{p3po}, where each task is trained with a variety of object sizes, most of our tasks are trained on a single object instance. Despite this limited diversity in the training data, \method{} demonstrates robust generalization capabilities. Figure~\ref{fig:rollouts_variations} depicts rollouts of \method{} for novel object instances. For a visual reference of the novel object instances used for each task, please refer to Appendix~\ref{appendix:novel_objects}. These results affirm \method{}'s strong generalization capabilities, making it suitable for real-world applications where encountering unseen objects is common.

\subsection{Can \method{} handle background distractors?}
\label{subsec:distractor}
We evaluate the robustness of \method{} in the presence of background clutter, as shown in Table~\ref{table:distractors}. This study is conducted on three tasks - \textit{put bread on plate}, \textit{sweep broom}, and \textit{put bottle on rack}. Trials are conducted using both in-domain and novel object instances. Examples of the distractors used are illustrated in Figure~\ref{fig:method}, with Figure~\ref{fig:rollouts_variations} depicting rollouts of \method{} in the presence of background distractors.  We observe that \method{} is robust to background clutter, exhibiting either comparable performance or only minimal degradation in the presence of background distractors. This robustness can be attributed to \method{}'s use of point-based representations, which are decoupled from raw pixel values. By focusing on semantically meaningful points rather than image-level features, \method{} enables policies that are resilient to environmental perturbations. \looseness=-1

\begin{table}[t]
\centering
\renewcommand{\tabcolsep}{4pt}
\renewcommand{\arraystretch}{1}
\caption{Policy performance of \method{} with teleoperated robot data on in-domain object instaces.}
\label{table:robot_data}
\begin{tabular}{ccccc}
\toprule
\multicolumn{1}{c}{\textbf{Demonstrations}} &
  \multicolumn{1}{c}{\textbf{\begin{tabular}[c]{@{}c@{}}Put bread \\ on plate\end{tabular}}} &
  \multicolumn{1}{c}{\textbf{\begin{tabular}[c]{@{}c@{}}Fold \\ towel\end{tabular}}} &
  \multicolumn{1}{c}{\textbf{\begin{tabular}[c]{@{}c@{}}Sweep \\ broom\end{tabular}}} &
  \multicolumn{1}{c}{\textbf{\begin{tabular}[c]{@{}c@{}}Make bottle \\ upright\end{tabular}}} \\ \midrule
Human            & 19/20 & \textbf{9/10} & \textbf{9/10} & \textbf{16/20} \\
Robot            & 18/20 & \textbf{9/10} & 4/10 & 12/20 \\
Human + Robot    & \textbf{20/20} & \textbf{9/10} & 8/10 & 8/20  \\ \bottomrule
\end{tabular}
\vspace{-1em}
\end{table}

\subsection{Can \method{} be improved with robot demonstrations?}
\label{subsec:robot_data}
Table~\ref{table:robot_data} investigates whether \method{}'s performance can be enhanced through co-training with teleoperated robot data, collected using a VR-based teleoperation framework~\cite{openteach}. We conduct this study on four tasks - \textit{put bread on plate}, \textit{fold towel}, \textit{sweep broom}, and \textit{make bottle upright}. For each task, we collect an equal number of robot demonstrations as human demonstrations, resulting in 30, 20, 20, and 30 demonstrations respectively. Interestingly, our findings reveal that for tasks involving complex motions, such as \textit{sweep broom} and \textit{make bottle upright}, policies trained solely on robot data perform poorly with the same amount of data as compared to those trained exclusively on human data. This drop in performance stems from the complex motions in these tasks making it harder to collect robot data using VR teleoperation, resulting in noisy demos. These results highlight an important consideration: humans and robots may execute the same task in different ways. Consequently, co-training with both human and robot data requires the development of algorithms capable of dealing with these differences effectively.

\begin{table}[t]
\centering
\renewcommand{\tabcolsep}{4pt}
\renewcommand{\arraystretch}{1}
\caption{The effect of triangulated depth on P3PO and \method{}.}
\label{table:depth}
\begin{tabular}{lccc}
\toprule
\multicolumn{1}{c}{\textbf{Method}} &
\multicolumn{1}{c}{\textbf{\begin{tabular}[c]{@{}c@{}}Put bread \\ on plate\end{tabular}}} &
\multicolumn{1}{c}{\textbf{\begin{tabular}[c]{@{}c@{}}Sweep \\ broom\end{tabular}}} & 
\multicolumn{1}{c}{\textbf{\begin{tabular}[c]{@{}c@{}}Put bottle \\ on rack\end{tabular}}} \\
\midrule
P3PO      & 0/20  & 0/10 & 0/30  \\
P3PO + Triangulated Depth & 17/20 & 4/10 & 23/30  \\
\method{} & \textbf{19/20} & \textbf{9/10} & \textbf{26/30}  \\
\method{} - Triangulated Depth & 0/20  & 0/10 & 0/30  \\
\bottomrule
\end{tabular}
\vspace{-1em}
\end{table}

\subsection{What design choices matter for human-to-robot learning?}
\label{subsec:design_considerations}

This section examines the impact of key design decisions on learning from human videos.

\paragraph{Depth Sensing} In \method{}, we utilize point triangulation from two camera views to obtain 3D key points, rather than relying on depth maps from the camera. We hypothesize that noisy camera depth leads to imprecise 3D key points, resulting in unreliable actions. Table~\ref{table:depth} tests this hypothesis on 4 real-world tasks by comparing the performance of P3PO and \method{} with and without triangulated depth. We observe that adding triangulated depth to P3PO improves its performance from 0\% to 72\%. Further, removing triangulated depth from \method{} reduces its performance from 90\% to 0\%. These results emphasize the importance of obtaining accurate 3D key points from human hands when learning robot policies from human videos. Appendix~\ref{appendix:depth_discrepancy} includes an illustration of imprecise actions resulting from noisy sensor depth. \looseness=-1

\begin{table}[t]
\centering
\renewcommand{\tabcolsep}{4pt}
\renewcommand{\arraystretch}{1}
\caption{Importance of object point inputs for policy learning.}
\label{table:object_pts}
\begin{tabular}{lcccc}
\toprule
\multicolumn{1}{c}{\textbf{Method}} &
\multicolumn{1}{c}{\textbf{\begin{tabular}[c]{@{}c@{}}Close \\ drawer\end{tabular}}} &
\multicolumn{1}{c}{\textbf{\begin{tabular}[c]{@{}c@{}}Put bread \\ on plate\end{tabular}}} &
\multicolumn{1}{c}{\textbf{\begin{tabular}[c]{@{}c@{}}Fold \\ towel\end{tabular}}} &
\multicolumn{1}{c}{\textbf{\begin{tabular}[c]{@{}c@{}}Make bottle \\ upright\end{tabular}}} \\
\midrule
MT-$\pi$   & 2/10  & 2/20  & 0/10 & 0/20  \\
MT-$\pi$ + object points & 8/10  & 1/20  & 6/10 & 2/20  \\
\method{}  & \textbf{10/10} & \textbf{19/20} & \textbf{9/10} & \textbf{16/20}  \\
\bottomrule
\end{tabular}
\vspace{-1em}
\end{table}

\paragraph{Significance of Object Points} While \method{} uses robot and object key points as input to the policy, MT-$\pi$~\cite{motiontracks}, the best-performing baseline in Table~\ref{table:same_distribution}, only uses robot key points and obtains information about the rest of the scene through an input image. We hypothesize that using object points can improve policy learning performance, especially when there is a morphology gap between data collection and inference. Table~\ref{table:object_pts} tests this hypothesis by providing object points in addition to the robot points already passed as input into MT-$\pi$. We observe that adding object points improves the performance of MT-$\pi$ on select tasks(comprehensive results on all tasks included in Appendix~\ref{appendix:object_pts}), suggesting that including object points in the input offers a potential advantage. Nevertheless, \method{} outperforms both methods by 68\% across all tasks, emphasizing the efficacy of predicting 3D key points rather than 2D key points in image space.

%% file: documents/limitations.tex
In this work, we presented \method{}, a framework that enables learning robot policies exclusively from human videos, does not require real-world online interactions, and exhibits generalization to spatial variations, new object instances, and robustness to background clutter. 

\textbf{Limitations:} We recognize a few limitations in this work: $(1)$ \method{}'s reliance on existing vision models makes it susceptible to their failures. For instance, failures in hand pose detection or point tracking under occlusion have a detrimental effect on performance. However, with continued advances in computer vision, we believe that frameworks such as \method{} will become stronger over time. $(2)$ Point-based abstractions enhance generalization capabilities, but sacrifice valuable scene context information, which is crucial for navigating through cluttered or obstacle-rich environments. Future research focusing on developing algorithms that preserve sparse contextual cues in addition to the point abstractions in \method{} might help address this. $(3)$ While all our experiments are from a fixed third-person camera view, a large portion of human task videos on the internet are from an egocentric view ~\cite{ego4d, hoi4d}. Extending \method{} to egocentric camera views can help us utilize these vast repositories of human videos readily available on the internet.

%% file: documents/acknowledgements.tex
We would like to thank Enes Erciyes, Raunaq Bhirangi, and Venkatesh Pattabiraman for help with setting up the Franka robot and Nur Muhammad Shafiullah, Raunaq Bhirangi, Gaoyue Zhou, Lisa Kondrich, and Ajay Mandlekar for their valuable feedback on the paper. This work was supported by grants from Honda, Hyundai, NSF award 2339096, and ONR award N00014-22-1-2773. LP is supported by the Packard Fellowship.

%% file: documents/appendix.tex
\SH{Add model hyperparams, table of demo numbers, check the explanation of point triangulation, how is depth normalized}

\subsection{Background}
\label{appendix:background}

\subsubsection{Semantic Correspondence}
Finding corresponding points across multiple images of the same scene is a well-established problem in computer vision~\cite{sift, zitova2003image}. Correspondence is essential for solving a range of larger challenges, including 3D reconstruction~\cite{nerf,gs}, motion tracking~\cite{cotracker,pips,pips_plus,tapir}, image registration~\cite{zitova2003image}, and object recognition~\cite{segmentanything}. In contrast, semantic correspondence focuses on matching points between a source image and an image of a different scene (e.g., identifying the left eye of a cat in relation to the left eye of a dog). Traditional correspondence methods~\cite{zitova2003image, sift} often struggle with semantic correspondence due to the substantial differences in features between the images. Recent advancements in semantic correspondence utilize deep learning and dense correspondence techniques to enhance robustness~\cite{fu2020deep, huang2022learning, asic} across variations in background, lighting, and camera perspectives. In this work, we adopt a diffusion-based point correspondence model, DIFT~\cite{dift}, to establish correspondences between a reference and an observed image, which is illustrated in Figure~\ref{fig:correspondence}.

\subsubsection{Point Tracking}
Point tracking across videos is a problem in computer vision, where a set of reference points are given in the first frame of the video, and the task is to track these points across multiple frames of the video sequence. Point tracking has proven crucial for many applications, including motion analysis~\cite{aggarwal1999human}, object tracking~\cite{yilmaz2006object}, and visual odometry~\cite{nister2004visual}. The goal is to establish reliable correspondences between points in one frame and their counterparts in subsequent frames, despite challenges such as changes in illumination, occlusions, and camera motion. While traditional point tracking methods rely on detecting local features in images, more recent advancements leverage deep learning and dense correspondence methods to improve robustness and accuracy~\cite{cotracker, pips, pips_plus}. In this work, we use Co-Tracker~\cite{cotracker} to track a set of reference points defined in the first frame of a robot's trajectory. These points tracked through the entire trajectory are then used to train generalizable robot policies for the  real world. 

\subsection{Algorithmic Details}
\label{appendix:alg_details}

\subsubsection{Point Triangulation}
\label{appendix:point_triangulation}


Point triangulation is a fundamental technique in computer vision used to reconstruct 3D points from their 2D projections in multiple images. Given \(n\) cameras with known projection matrices \(P_1, P_2, ..., P_n\) and corresponding 2D image points \(x_1, x_2, ..., x_n\), the goal is to find the 3D point \(X\) that best explains these observations.

The projection of \(X\) onto each image is given by:

\[ x_i \sim P_i X \]

where \(\sim\) denotes equality up to scale.

One common approach is the Direct Linear Transform (DLT) method:

\begin{enumerate}
    \item For each view \(i\), we can form two linear equations:
    \[ x_i(p^3_i \cdot X) - (p^1_i \cdot X) = 0 \]
    \[ y_i(p^3_i \cdot X) - (p^2_i \cdot X) = 0 \]
    where \(p^j_i\) is the \(j\)-th row of \(P_i\).
    
    \item Combining equations from all views, we get a system \(AX = 0\).
    
    \item The solution is the unit vector corresponding to the smallest singular value of \(A\), found via Singular Value Decomposition (SVD).
\end{enumerate}

For optimal triangulation, we aim to minimize the geometric reprojection error. 

\subsection{Hyperparameters}
\label{appendix:subsec:hypperparams}
The complete list of hyperparameters is provided in Table~\ref{table:hyperparams}. Details about the number of demonstrations for each task has been included in Section~\ref{subsec:task_desc}, and summarized in Table~\ref{table:num_demos}. All the models have been trained using a single NVIDIA RTX A4000 GPU.

\begin{table*}[!h]
    \caption{List of hyperparameters.}
    \label{table:hyperparams}
    \begin{center}
    \setlength{\tabcolsep}{18pt}
    \renewcommand{\arraystretch}{1.5}
    \begin{tabular}{ l l } 
        \toprule
        Parameter                  & Value \\
        \midrule
        Learning rate              & $1e^{-4}$\\
        Image size                 & $256\times 256$ (for BC, BC w/ Depth, MT-$\pi$)\\
        Batch size                 & 64 \\
        Optimizer                  & Adam\\
        Number of training steps   & 100000 \\
        Transformer architecture   & minGPT~\cite{minGPT} (for BC, BC w/ Depth, P3PO, \method{}) \\
                                   & Diffusion Transformer~\cite{track2act} (for MT-$\pi$) \\
        Hidden dim                 & 256\\
        Observation history length & 1 (for BC, BC w/ Depth) \\
                                   & 10 (for MT-$\pi$, P3PO, \method{}) \\
        Action head                & MLP \\
        Action chunk length        & 20\\ \bottomrule
    \end{tabular}
    \end{center}
\end{table*}

\begin{table*}[!h]
    \caption{Number of demonstrations.}
    \label{table:num_demos}
    \begin{center}
    \setlength{\tabcolsep}{18pt}
    \renewcommand{\arraystretch}{1.5}
    \begin{tabular}{ l c c } 
        \toprule
        Task                    & Number of object instances  & Total number of demonstrations \\
        \midrule
        Close drawer            & 1   & 20 \\
        Put bread on plate      & 1   & 30 \\
        Fold towel              & 1   & 20 \\
        Close oven              & 1   & 20 \\
        Sweep broom             & 1   & 20 \\
        Put bottle on rack      & 2   & 30 \\
        Put bowl in oven        & 1   & 20 \\
        Make bottle upright     & 2   & 30 \\
        \bottomrule
    \end{tabular}
    \end{center}
\end{table*}




\subsection{Implementation Details for MT-$\pi$}
\label{appendix:motion_tracks}
Since the official implementation of MT-$\pi$ is not yet public available, we adopt the Diffusion Transformer (DiT) based implementation of a 2D point track prediction model proposed by \citet{track2act}. We modify the architecture such that given a single image observation and robot motion tracks on the image, the model predicts future tracks of the robot points. These robot tracks are then converted to 3D using corresponding tracks for two camera views. The robot action is then computed from the 3D robot tracks using the same rigid-body geometry constraints as \method{} (described in Section~\ref{subsec:robot_action_from_points}). MT-$\pi$ proposes the use of a key point retargeting network in order to convert the human hand and robot key points to the same space. Since we already convert the human hand key points to the corresponding robot points for \method{}, we directly use these converted robot points instead of learning a separate keypoint retargeting network.

To ensure the correctness of our implementation, we evaluate MT-$\pi$ in a setting identical to the one described in their paper. We conduct this evaluation on the \textit{put bread on plate} task. We use 30 robot teleoperated demonstrations in addition to the human demonstrations, resulting in a total of 60 demonstrations. We observed a performance of 18/20, thus, confirming the correctness of the implementation.

\subsection{Experiments}
\label{appendix:experiments}

\subsubsection{Illustration of Spatial Generalization and Novel Object Instances}
\label{appendix:novel_objects}

\begin{figure*}[th]
\centering
\includegraphics[width=0.9\linewidth]{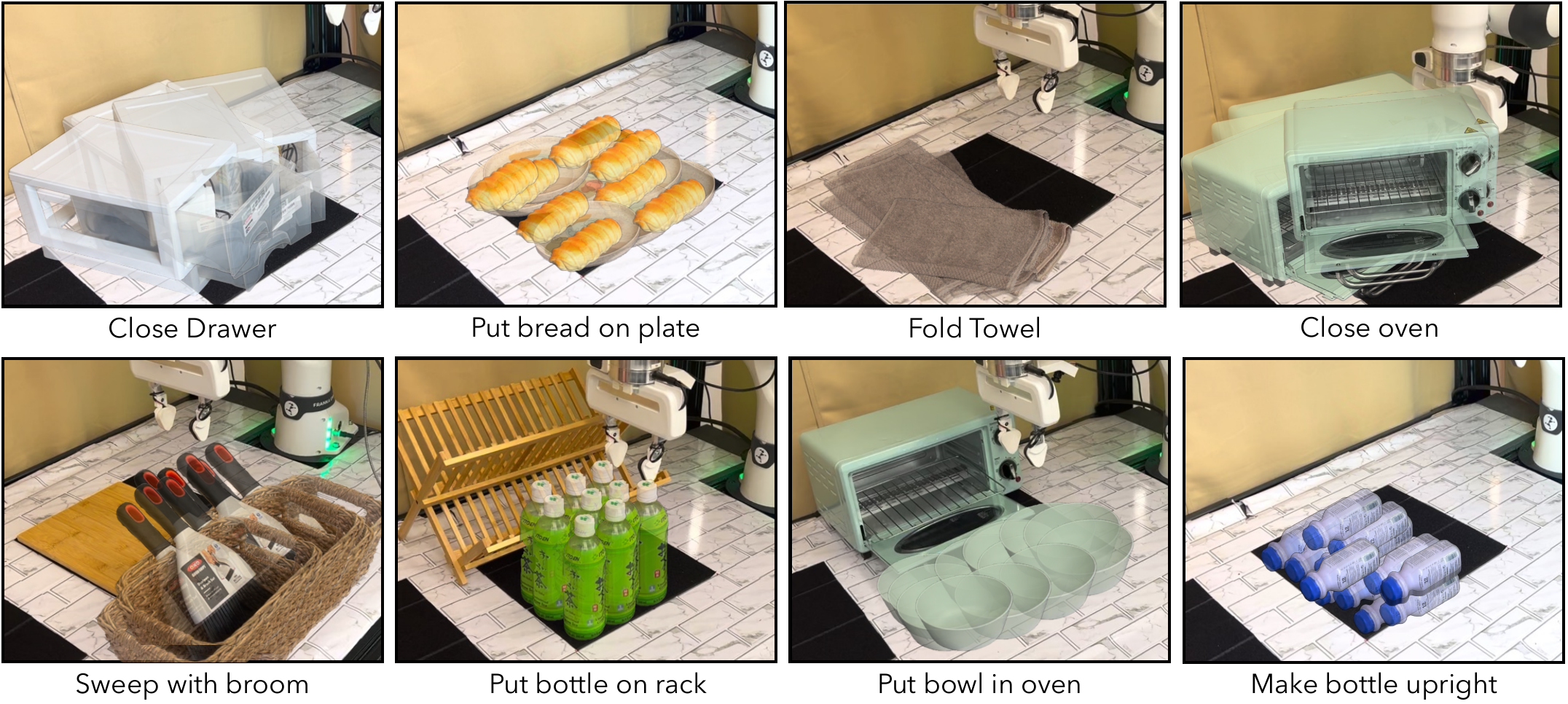}
\caption{ Illustration of spatial variation used in our experiments.}
\label{fig:spatial_gen_all}
\end{figure*}

\begin{figure*}[th]
\centering
\includegraphics[width=0.65\linewidth]{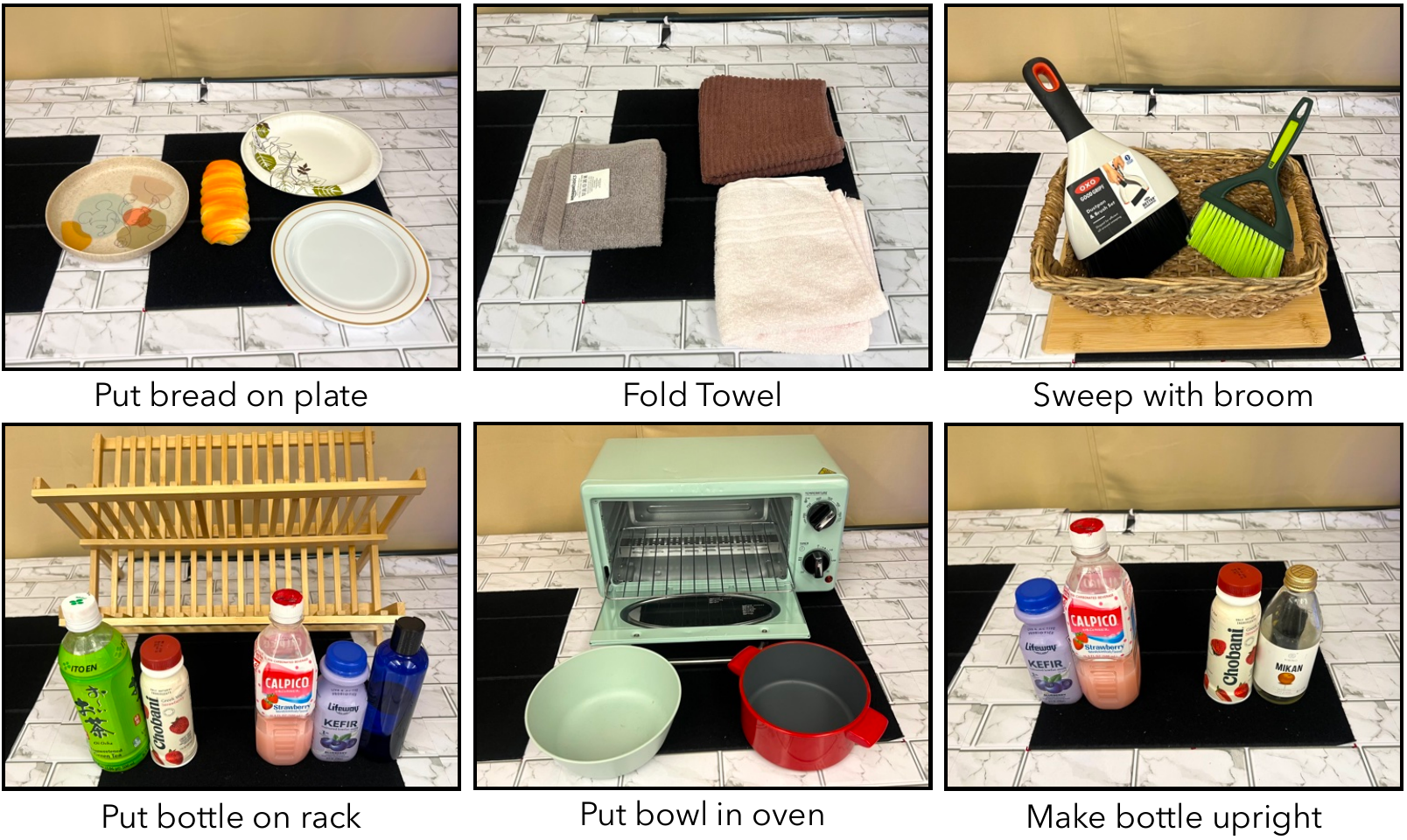}
\caption{Illustration of objects used in our experiments. For each task, on the left are in-domain objects while on the right are novel objects used in our generalization experiments.}
\label{fig:novel_objects_all}
\end{figure*}

Figure~\ref{fig:spatial_gen_all} and Figure~\ref{fig:novel_objects_all} illustrate the variations in object positions and novel object instances used for each task, respectively.

\begin{figure*}[t]
\centering
\includegraphics[width=0.9\linewidth]{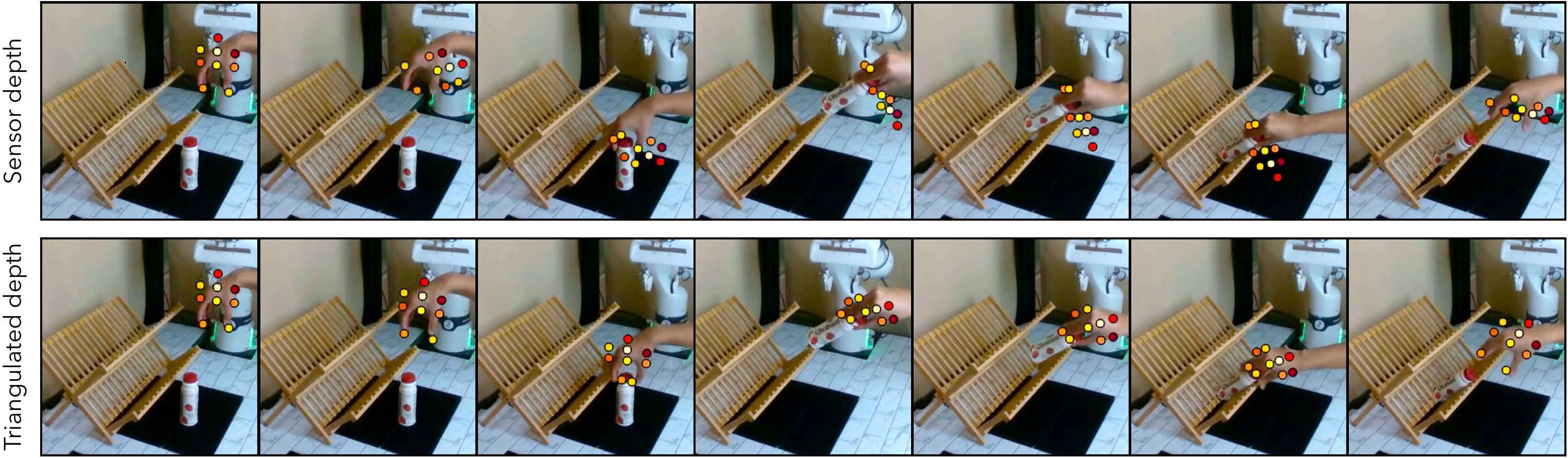}
\caption{ Illustration of discrepancy in actions obtained from sensor depth and triangulated depth for the task of putting a bottle on the rack.}
\label{fig:depth}
\end{figure*}

\subsubsection{Illustration of Depth Discrepancy}
\label{appendix:depth_discrepancy}
Figure~\ref{fig:depth} provides an illustration of the discrepancy in actions obtained from sensor depth and triangulated depth for the task of putting a bottle on the rack. We observe that the noise in sensor depth leads to noise in robot points which is turn results in unreliable actions.

\subsubsection{Significance of Object Points}
\label{appendix:object_pts}
Table~\ref{table:object_pts_in_domain} and Table~\ref{table:object_pts_new_objects} study the performance of MT-$\pi$ with and without object points and \method{} across all of our tasks. We observe that MT-$\pi$ with object points outperforms MT-$\pi$ on select tasks, suggesting that including object points in the input offers a potential advantage.

\begin{table*}[t]
\centering
\caption{In-domain policy performance}
\label{table:object_pts_in_domain}
\renewcommand{\tabcolsep}{4pt}
\renewcommand{\arraystretch}{1}
\begin{tabular}{lcccccccc}
\toprule
\multicolumn{1}{c}{\textbf{Method}} & \textbf{\begin{tabular}[c]{@{}c@{}}Close\\ drawer\end{tabular}} & \textbf{\begin{tabular}[c]{@{}c@{}}Put bread \\ on plate\end{tabular}} & \textbf{\begin{tabular}[c]{@{}c@{}}Fold\\ towel\end{tabular}}& \textbf{\begin{tabular}[c]{@{}c@{}}Close\\ oven\end{tabular}} & \textbf{\begin{tabular}[c]{@{}c@{}}Sweep\\ broom\end{tabular}} & \textbf{\begin{tabular}[c]{@{}c@{}}Put bottle\\ on rack\end{tabular}} & \textbf{\begin{tabular}[c]{@{}c@{}}Put bowl\\ in oven\end{tabular}} & \textbf{\begin{tabular}[c]{@{}c@{}}Make bottle\\ upright\end{tabular}} \\ \bottomrule
MT-$\pi$~\cite{motiontracks}      & 2/10 & 2/20 & 0/10 & 4/10 & 0/10 & 8/30 & 0/10 & 0/20 \\
MT-$\pi$ + object points & 1/20 & 6/10 & 1/20 & 4/10 & 0/10 & 0/10 & 2/20 & 8/10 \\
\hdashline\noalign{\vskip 0.5ex}
\method{} \textbf{(Ours)}         & \textbf{10/10} & \textbf{19/20} & \textbf{9/10} & \textbf{9/10} & \textbf{9/10} & \textbf{26/30} & \textbf{8/10} & \textbf{16/20} \\ \bottomrule
\end{tabular}
\end{table*}

\begin{table*}[t]
\centering
\renewcommand{\tabcolsep}{4pt}
\renewcommand{\arraystretch}{1}
\caption{Policy performance on novel object instances}
\label{table:object_pts_new_objects}
\begin{tabular}{lcccccc}
\toprule
\multicolumn{1}{c}{\textbf{Method}} & \textbf{\begin{tabular}[c]{@{}c@{}}Put bread \\ on plate\end{tabular}} & \textbf{\begin{tabular}[c]{@{}c@{}}Fold\\ towel\end{tabular}} & \textbf{\begin{tabular}[c]{@{}c@{}}Sweep\\ broom\end{tabular}} &  \textbf{\begin{tabular}[c]{@{}c@{}}Put bottle\\ on rack\end{tabular}} & \textbf{\begin{tabular}[c]{@{}c@{}}Put bowl\\ in oven\end{tabular}} & \textbf{\begin{tabular}[c]{@{}c@{}}Make bottle\\ upright\end{tabular}} \\ \midrule
MT-$\pi$~\cite{motiontracks} & 1/20 & 0/20 & 0/10 & 0/30 & 0/10 & 0/20 \\
MT-$\pi$ + object points & 2/20 & 0/20 & 0/20 & 1/10 & 0/10 & 1/20 \\
\hdashline\noalign{\vskip 0.5ex}
\method{} \textbf{(Ours)}    & \textbf{18/20} & \textbf{15/20} & \textbf{4/10} & \textbf{27/30} & \textbf{9/10} & \textbf{9/20} \\ \bottomrule
\end{tabular}
\end{table*}